\documentclass[letterpaper, 10 pt, conference]{ieeeconf}  
\IEEEoverridecommandlockouts                              

\usepackage{microtype}
\usepackage{comment}

\usepackage{graphicx} 
\graphicspath{{./figures/}}
\usepackage{subcaption}
\usepackage{float}

\usepackage{amsmath} 
\usepackage{amssymb}  
\usepackage{units}
\usepackage{bm} 
\usepackage{multirow}
\usepackage{todonotes}

\usepackage{algorithm}
\usepackage{algpseudocode}

\makeatletter
\let\NAT@parse\undefined
\makeatother

\usepackage[square, numbers, sort&compress]{natbib}
\usepackage[hidelinks]{hyperref}
\usepackage{cleveref}

\usepackage{graphicx} 
\graphicspath{{./figures/}}
\usepackage{subcaption}
\usepackage[font=small]{caption}
\usepackage{xparse}
\usepackage{xcolor}
\usepackage{tabularx}
\usepackage{diagbox}
\usepackage{makecell}

\title{\LARGE \bf Learning Dynamic Bipedal Walking Across Stepping Stones}

\author{
Helei Duan, Ashish Malik, Mohitvishnu S. Gadde, Jeremy Dao, Alan Fern, Jonathan Hurst
\thanks{*This work is supported by the NSF Grant No. IIS-1849343 and DARPA Contract W911NF-16-1-0002.}
\thanks{All authors are with Collaborative Robotics and Intelligent Systems Institute, Oregon State University, Corvallis, Oregon, 97331, USA. Email: \{{\tt\footnotesize duanh, malikas, gaddem, daoje, afern, jhurst}\}@oregonstate.edu. }
}

\begin{document}

\maketitle
\thispagestyle{empty}
\pagestyle{empty}

\begin{abstract}

In this work, we propose a learning approach for 3D dynamic bipedal walking when footsteps are constrained to stepping stones. While recent work has shown progress on this problem, real-world demonstrations have been limited to relatively simple open-loop, perception-free scenarios. Our main contribution is a more advanced learning approach that enables real-world demonstrations, using the Cassie robot, of closed-loop dynamic walking over moderately difficult stepping-stone patterns. Our approach first uses reinforcement learning (RL) in simulation to train a controller that maps footstep commands onto joint actions without any reference motion information. We then learn a model of that controller's capabilities, which enables prediction of feasible footsteps given the robot's current dynamic state. The resulting controller and model are then integrated with a real-time overhead camera system for detecting stepping stone locations. For evaluation, we develop a benchmark set of stepping stone patterns, which are used to test performance in both simulation and the real world. Overall, we demonstrate that sim-to-real learning is extremely promising for enabling dynamic locomotion over stepping stones. We also identify challenges remaining that motivate important future research directions.

\end{abstract}
\section{Introduction}
Bipedal robots have advantages over wheeled and quadruped robots in many environments, especially those designed for human locomotion. However, current bipedal systems are still far from reaching animal levels of agility and robustness, especially over highly constrained discrete terrains such as stepping-stone-like environments. 
In order to traverse such environments, the robot must reason both about its own dynamics and the surrounding terrain to determine achievable footstep locations while maintaining balance\cite{Matthis2018}. 
The primary goal of this work is to study a learning-based approach to this problem that learns a controller for achieving specified footsteps and a model of the controller's reachable footsteps given the robot state.


\begin{figure}[!ht]
\centering
  \includegraphics[width=0.99\columnwidth,angle=0]{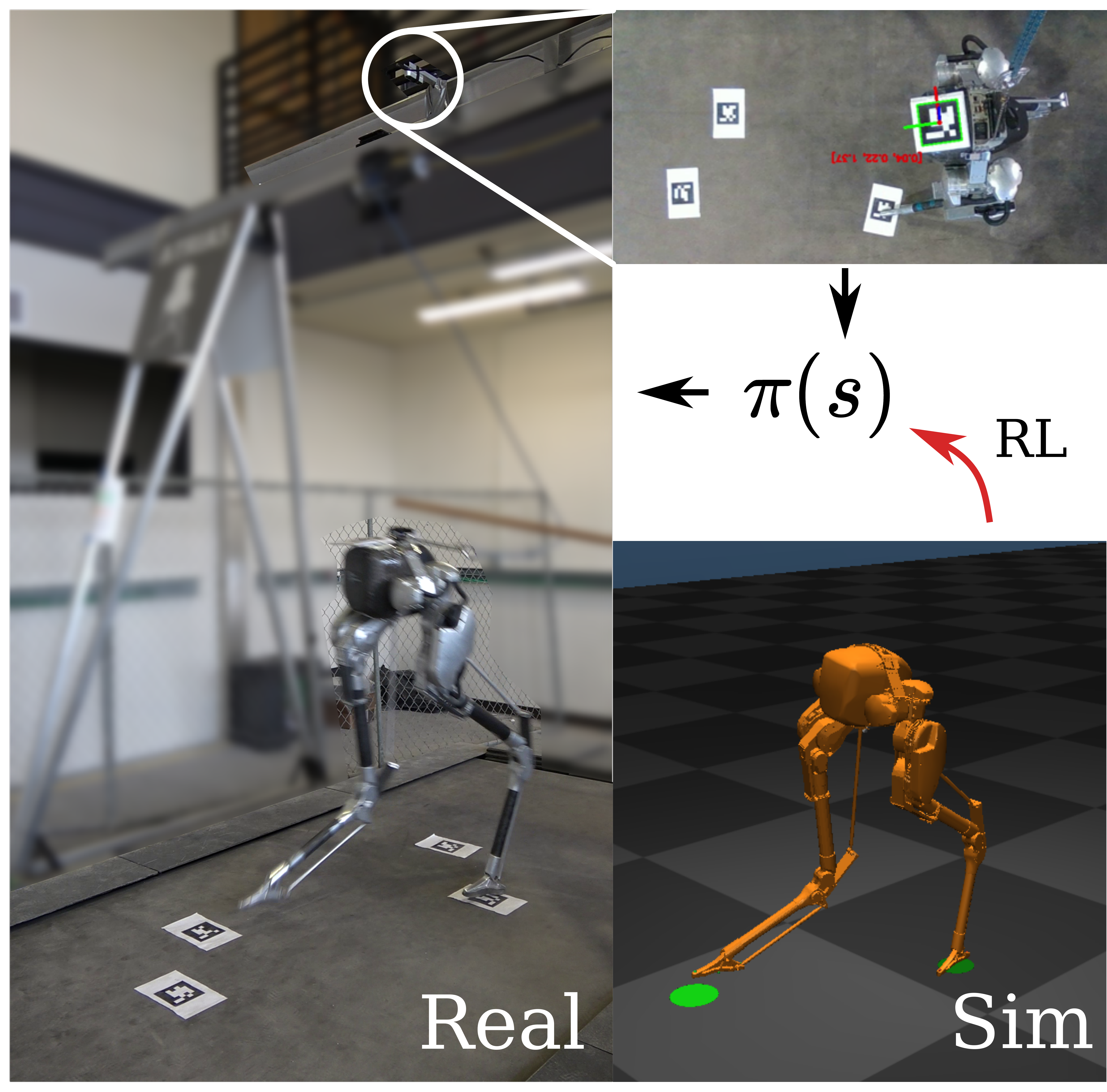}
  \caption{The robot Cassie only knows the immediate next footstep target, and is reaching to the next marker. 
  A fixed camera is placed overhead of the treadmill area to provide a real-time estimation of Cassie's position relative to the next footstep target shown as the markers on treadmill.
  Green dots in simulation are the footstep targets. 
  Cassie performs the same stepping stone pattern between hardware and simulation.}
  \label{fig:lead}
\end{figure}


Dynamic bipedal walking over discrete terrains requires controllers to deliberately target each footstep, while traveling at a reasonable speed and maintaining balance. Several previous works focus on using model-based techniques to design controllers, such as using an offline gait library and online interpolation \cite{Nguyen2017a, Nguyen2020}, a control barrier function \cite{7798370}, or reduced-order models \cite{Dai2021}. These techniques often involve a sequence of optimizations that must be computed online.

Reinforcement learning (RL) approaches have the potential to learn a fast online controller via large amounts of offline training in simulation. Sim-to-real RL has recently demonstrated highly-dynamic bipedal gaits on hardware without explicit constraints on footstep locations \cite{Castillo2021, bipedalgaitsICRA, Krishna2022}. RL has also recently produced controllers that enable legged robots to walk over discrete terrains, e.g. using curriculum-based training of bipeds for simulation only \cite{Xie2020a} or hierarchical control structures for quadrupeds \cite{Xie2021, Yu2021, pmlr-v164-margolis22a, deepgait, Gangapurwala2020}. These techniques, however, have yet to demonstrate real-world bipedal dynamic walking over stepping stones. Most recently, sim-to-real RL has demonstrated bipedal walking for open-loop footstep constraints that are known before deployment \cite{Duan2022}. However, there is currently no RL-based approach that has demonstrated closed-loop control of bipedal dynamic walking over stepping stones.



In this paper, we fill this empirical gap by significantly extending prior work \cite{Duan2022} in order to produce real-world demonstrations of a closed-loop system. Our specific contributions include: 
\begin{itemize}
    \item We propose a new learning-based control architecture for dynamic walking stepping stone policies that control both the motor actuation and the step frequency. Further this new architecture supports bootstrapping from previously learned controllers for unconstrained locomotion.
    \item We learn a predictive model that encodes raw observations into a latent dynamics space where reachable footstep locations can be predicted for lookahead. 
    \item We demonstrate sim-to-real learning of the controller and predictive model that are evaluated both in simulation and on the real robot Cassie. 
    \item We propose an evaluation approach in terms of a set of benchmark stepping stone patterns that can be used and extended in future research to track progress. 
\end{itemize}

\section{Sim-to-Real Learning for Stepping Stones}
Here, we discuss the architecture of the control policy and the corresponding sim-to-real RL training approach.  

\subsection{Control Policy Design}



We follow common terminology used in RL where the goal is to learn a reward-maximizing control policy that takes states (or observations) as input and produces actions as output. In this paper the main objective of the control policy is to allow the robot to reach the next footstep target while maintaining balance. Below we describe the state space, action space, and overall architecture of the policy used in this work. 

\subsubsection{State Space} 
The input state to the policy includes: 1) the robot proprioceptive states, providing the position and velocity of each motor (10 motors in total) as well as the body orientations and angular velocities, 2) a periodic clock value $\phi$ that is reset to zero when the period ends, and 3) the footstep command for the next step, given as the relative position from the base frame of the pelvis to the target. In this work, footstep targets are assumed to be in the same vertical plane and hence are two-dimensional. Targets for the next step are updated based on the value of the periodic clock, which aligns with foot touchdown events as facilitated by the reward during training (see Section \ref{sec:reward}). Importantly, this ego-centric encoding of footstep commands avoids the need for real-time estimation of feet contact and positions.
%

\subsubsection{Action Space}
Following prior work \cite{Siekmann2021}, the RL policy operates at 40Hz and outputs PD set-points for all motors, which are provided to PD controllers operating at 2kHz. In addition, our RL policy outputs a clock increment $\Delta\phi$, which adjusts the clock value by $\phi_{i+1}=\phi_{i}+\Delta\phi$ at each control update. This choice is motivated by biomechanics studies \cite{Bertram2005, Matthis2018}, which show that the stepping frequency plays an important role in adjusting the system dynamics to robustly vary the step length.

In the most closely related prior work \cite{Duan2022}, hand-crafted heuristics were used to adjust the clock period, which is tedious since the choice of step frequency interacts with system dynamics and the overall learning process. Rather, in this work, our policy directly learns to jointly control the clock period and PD set-points in order to maximize overall reward. We limit the range of the clock period so that gait cycle times are constrained to be within [0.65, 1.2] seconds. Importantly, the clock period is updated at 40Hz, which allows the learned solution to adjust to sudden changes in step size. 

\subsubsection{Policy Architecture}
We use a neural network to represent the policy for mapping state sequences to actions. Training footstep-constrained policies from scratch using RL is computationally expensive, requiring hundreds of millions of simulated time steps. Thus, in this work, we introduce a novel neural network architecture for the policy that supports transfer from previously learned locomotion policies to speed up learning. A challenge is that previously learned locomotion policies will typically use different state and/or action spaces and be optimized for different reward functions. However, nearly all such policies will use the proprioceptive state as one of the inputs for encoding the dynamic state of the robot. Thus, our new architecture is divided into a module that solely encodes the proprioceptive state history, which can be transferred between policies.

\begin{figure}[t]
\vspace{1pt}
\centering
  \includegraphics[width=0.99\columnwidth,angle=0]{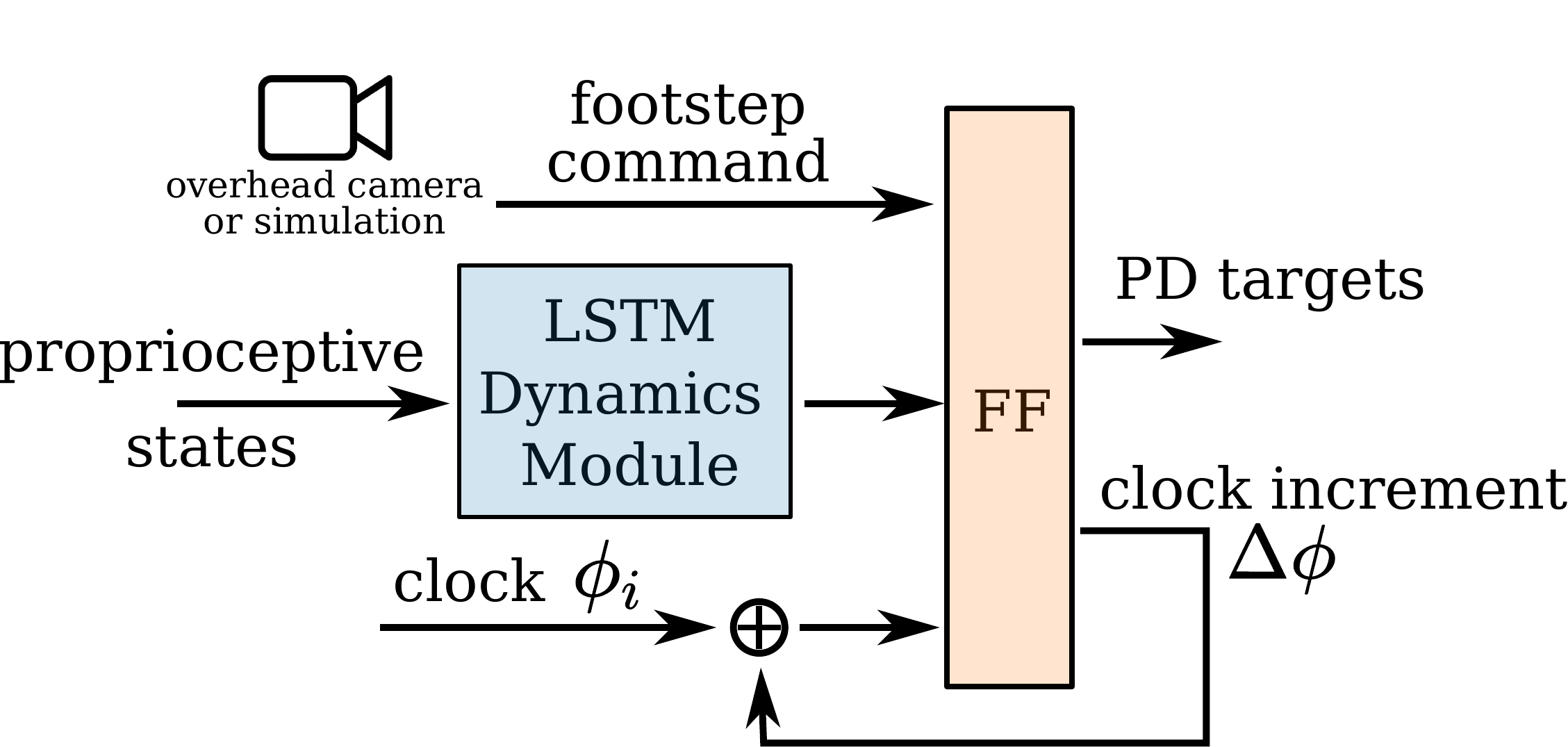}
  \caption{To allow for bootstrapping from pretrained policies while having different inputs, the policy network has a LSTM module which only consumes proprioceptive states. The feed-forward (FF) layer concatenates the output of the LSTM layers with a periodic clock and the commands and eventually outputs the motor actions. This means we can re-use the LSTM layers, because it is not affected by the task specific inputs like commands. }
  \label{fig:policy_arch}
\end{figure}

More specifically, we decompose the neural network policy into two parts as shown in Figure \ref{fig:policy_arch}: one to handle the robot's proprioceptive inputs as an LSTM dynamics module, and another feed-forward (FF) module to handle various types of command inputs that may differ among tasks. For our stepping stone application, the feed-forward module outputs motor set-points along with the clock increment $\Delta\phi$. 
Intuitively, the dynamics module is intended to provide a generic, but useful, representation of the robot's dynamic state based on the history of observations, which is useful for a variety of tasks. Our proposed architecture allows for initializing our dynamics module with modules learned from previous tasks, even if the previous tasks required different command inputs. This is a form of transfer learning that is common in other areas of machine learning, such as computer vision where pre-trained models are commonly used as starting points for learning. Our experiments show that this type of transfer can significantly speedup learning of stepping-stone locomotion.


\subsection{Training in Simulation} 

We follow a sim-to-real RL training approach similar to the the most closely related prior work \cite{Duan2022}. Below we provide an overview of the approach and refer the reader to that prior work for more specific training details. In addition, we describe our transfer learning approach with is a new addition to this work.

\subsubsection{Training Scenario Generation}

All training is conducted in simulation using the Proximal Policy Optimization (PPO) \cite{Schulman2017} RL algorithm. To help support reliable sim-to-real transfer we use the same dynamics randomization technique of prior work \cite{Siekmann-RSS-20}. In particular, each training episode uses a set of randomized physical parameters (e.g. friction, center of mass, etc.) drawn uniformly from a range of values, which encourages the learned policies to be robust to those variations.

For each training episode, we randomly pre-generate a sequence of relative footstep commands to ensure that the net movement of the robot over an episode follows meaningful walking directions, such as forward, backward or stepping in place. During training, we input commands from the sequence in order. This supports training on a wide variety of footstep commands from a wide variety of dynamic states.

\subsubsection{Reward Function}
\label{sec:reward}

We use the same reward function as prior work \cite{Duan2022}, which depends on the periodic clock, the robot state, and the command input. The reward encourages alternating swing and stance phases to align with the clock as well as reaching the next footstep target, which is an input to the policy. This training setup encourages the robot to trade-off between maintaining balance and reaching to the footstep target. Thus, the footstep matching reward acts as a soft constraint that the robot may violate in favor of not falling down, in order to collect more reward through out the training episode.

\subsubsection{Transfer of Pre-trained Dynamics Module} 

Training the neural network policy from scratch for challenging locomotion tasks can require days of training on multiple processors. To test the potential speedup, we consider transferring a pre-trained dynamics model from a conventional dynamic walking policy. Specifically, we train a regular locomotion policy that is able to achieve speed and direction commands following prior work \cite{bipedalgaitsICRA}. The only adjustment is that we use the above policy architecture, which includes the dedicated dynamics module, but with the desired speed and direction used as command inputs rather than footstep targets. After fully training this policy, the dynamics module was saved for transfer. Next we trained the stepping-stone policy by initializing it with this pre-trained dynamics module. During training, PPO is used to adjust the parameters of the entire network, including fine-tuning the dynamics module. 

\section{Learning a Reachability Prediction Model}
\label{sec:model learning}

\begin{figure}
\vspace{4pt}
\centering
  \includegraphics[width=0.99\columnwidth,angle=0]{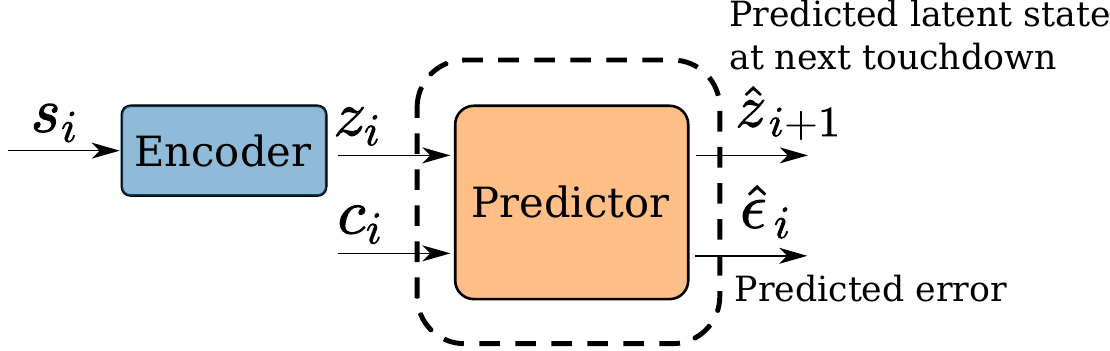}
  \caption{Structure of the reachability prediction model, where the encoder produces the latent dynamic states and the predictor can be called recursively.
}
  \label{fig:predict_arch}
\end{figure}

Given a trained stepping stone policy, 
the reachable touchdown locations that can be reliably achieved is a function of the robot's dynamic state. Higher-level planning for footstep selection needs a model of this function to achieve dependable performance. To this end, we learn a step error prediction model which we call the reachability model. The reachability model maintains a latent dynamic state of the robot, which is used to predict: 1) the step error of the current step target, and 2) the next latent robot state after touchdown of the current step attempt (Figure \ref{fig:predict_arch}). We use this model to deduce the reachable region as potential step targets where the predicted step-error is less than some threshold. Since the model also produces the next robot latent state, the model can also be used to direct multi-step look-ahead search for highly constrained terrains.


\subsection{Training Data Collection}
\label{sec:pred_data_collect}

During data collection, each episode first runs the policy by providing it a with a randomized footstep commands that is fixed for several steps, which we refer to as the procedural steps. After the procedural steps the robot state is recorded and used to generate multiple training examples. We then perform the collection step, where each training example is created by selecting one of the procedurally-generated robot states and then executing the policy with a randomized step target. The step error $\epsilon$ for each of the step targets is recorded as ground-truth values for training.  This process is repeated with the same procedural steps and different collection step commands to cover a full range of X-Y Cartesian space of footstep commands. Thus, each randomized procedural command embodies a range of collection step commands. For our purposes, we limit the the step target range to be within [0, 0.5]m in the X-direction and [-0.2, 0.2]m in the Y-direction for randomizing the procedural steps and the collection steps. The result is a supervised training set of robot states, footstep targets, and the error that resulted after attempting the step. Each training tuple has the form $(s_{i}, s_{i+1}, c_i, \epsilon_{i})$, where $s_i$ is the initial robot state, $c_i$ is the target step, $s_{i+1}$ is the robot state after touchdown, and $\epsilon_{i}$ is the touchdown error with respect to $c_i$.

We investigate different variations of the above data collection method to observe the impact on model accuracy.
The variations include: 1) Using dynamics randomization during data collection, 2) Adding a small amount of noise into the procedural step commands at each touchdown, 3) Averaging multiple step errors around the final procedural touchdown to account for contact uncertainties, and
4) Vanilla, without any additional scheme. We evaluate and compare each of these methods in Section \ref{sec:pred_model_results}. We collect separate evaluation data to test the models for a more candid comparison of the potential models instead of segregating each dataset into test and train sets. The evaluation data is collected by randomizing the step targets at each touchdown during the procedural steps.

\begin{figure*}[!t]
\vspace{4pt}
	\centering
    \includegraphics[width=0.88\linewidth]{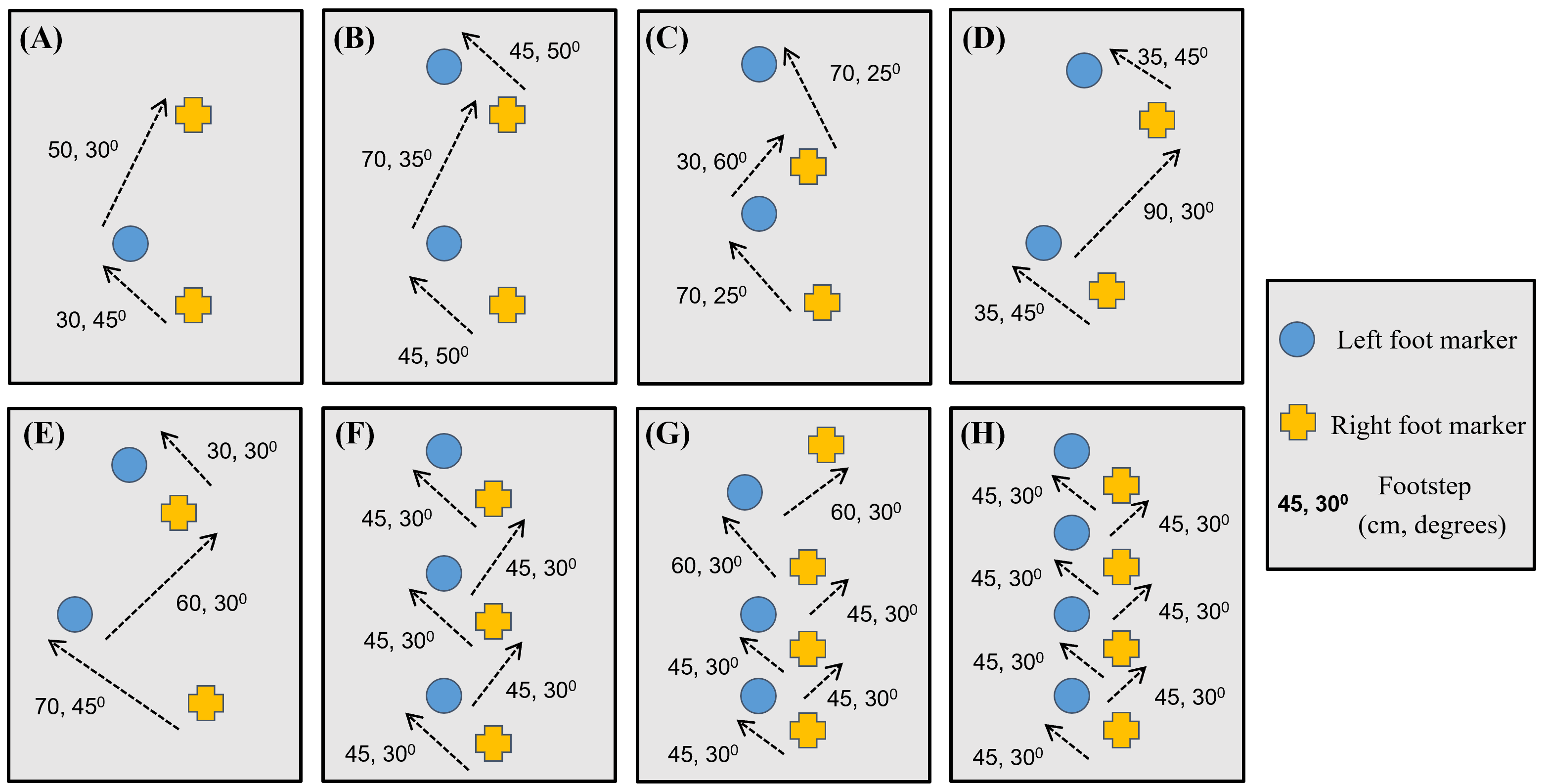}
    \caption{A set of stepping stone patterns that gradually change the number of required footsteps and the difficulty. The patterns are designed with step length and step direction, where blue/yellow symbols indicate left/right sides.}
    \label{fig:steppingstone_patterns}
    \vspace{-0.7cm}
\end{figure*}

\subsection{Training the Reachability Prediction Model}

The complete architecture of the reachability model is shown in Figure \ref{fig:predict_arch}. The encoder network learns a latent state representation that aims to capture information relevant to the ensuing touchdown error. The second network called the predictor uses this latent representation with step targets to predict the step error and the latent representation of the next touchdown state. Note that this architecture, in concept, supports auto-regressive prediction of multiple steps into the future by feeding predicted latent states into the predictor network.

We use feed-forward neural networks with ReLu activation functions for each module and train them end-to-end using supervised learning. For each training tuple $(s_{i}, s_{i+1}, c_i, \epsilon_{i})$ the network loss function is computed as follows: The current encoder is used to encode the robot states $s_i$ and $s_{i+1}$ to latent states $z_i$ and $z_{i+1}$ respectively. Then, the predictor takes in $z_i$ and $c_i$ and outputs the predicted errors for the next touchdown $\hat{\epsilon}_{i}$ and the prediction of the next latent state $\hat{z}_{i+1}$. 
The loss $L$ for the training example is then just a weighted error of each prediction. That is, 
\begin{align}
    L &= (\epsilon_{i} - \hat{\epsilon}_{i})^2 + \gamma (z_{i+1} - \hat{z}_{i+1})^2 \nonumber
\end{align}
where, $\gamma = 5$ is the weighing constant.

\section{Experiment Setup} 

This section explains our experimental setup for simulation and hardware. First, we describe a benchmark set of stepping-stone patterns patterns. Second, we describe our perception system, based on an overhead camera, for providing real-time information about footstep targets.  


\subsection{Stepping Stone Patterns for Evaluation}
Ideally an evaluation should enable multiple researchers to compare performance across varying robot platforms. To enable such comparisons, we designed a set of 8 stepping stone patterns of varying difficultly, which are illustrated in Figure \ref{fig:steppingstone_patterns}. Each pattern specifies a number of alternating left-right footsteps and specifies the distance and direction between successive steps. The set of patterns were designed to span different dimensions of difficulty. One dimension is the number of steps, which tests the ability to address error accumulation over successive step sequences. A second dimension is to vary the length and angle of steps, with more extreme values tending to be more difficult. Finally, a third dimension is to vary the relative length and/or angle between successive steps, e.g. a short step followed by a long step, which tends to increase difficulty with more abrupt changes.

In simulation we quantitatively evaluate policies on the patterns via the average and maximum step error as measured by the mid-point of the foot at touchdown relative to the center position of the footstep target. The maximum error is particularly informative, since it indicates the minimum size of a stepping stone that would have led to a successful traversal of a pattern. Such quantitative evaluation is more difficult on real hardware and we instead report visual assessments of performance in this paper.

Note that these patterns are designed for the robot to exactly follow the specified steps rather than needing to choose the best steps. We expect that future pattern benchmarks will be designed that also test the higher-level decision of stepping stone selection. In this work, we provide an initial evaluation of such decision making that is independent of our 8 stepping-stone patterns (Sections \ref{sec:targeting1} and \ref{sec:targeting2}).


\subsection{Overhead Camera for Target Footstep Estimation} 
In order to execute our stepping-stone policy on hardware, the robot needs real-time information about the next footstep target relative to the robot pelvis. Since the focus of this work is not on perception, we simplified this problem via the use of an overhead camera, rather than a camera on-board the robot. Specifically, we installed a downward facing Intel D435 depth camera positions above a treadmill area (Figure \ref{fig:lead}). Each footstep of a stepping-stone pattern was represented via an ArUco marker \cite{Romero-Ramirez2018, Garrido-Jurado2016} attached to the treadmill at the specified relative locations. We also attach a marker on top of the robot pelvis to track its current position and orientation. The RGB information provided by the camera is then used to track the marker center points, noting that the robot corresponds to the only moving marker. It is then straightforward to provide the relative location of the next footstep target to the policy at any point during a pattern traversal. In order to avoid having a cable attached to the robot, the camera is connected to an external compute platform and sends the information via wireless network to the robot.
\section{Simulation Results}
This section presents experimental results of the entire control system over various evaluations in simulation. 
First, we test the policy in simulation over the set of patterns. Second, we evaluate the accuracy and utility of the learned prediction model when there is more than one footstep choice given the nearby terrain. 

\subsection{Utility of Pretrained Dynamics Module}
\label{sec:sim_pretrain_results}
\begin{figure*}[t]
	\centering
    \begin{subfigure}{0.34\textwidth}
         \centering
         \includegraphics[width=\textwidth]{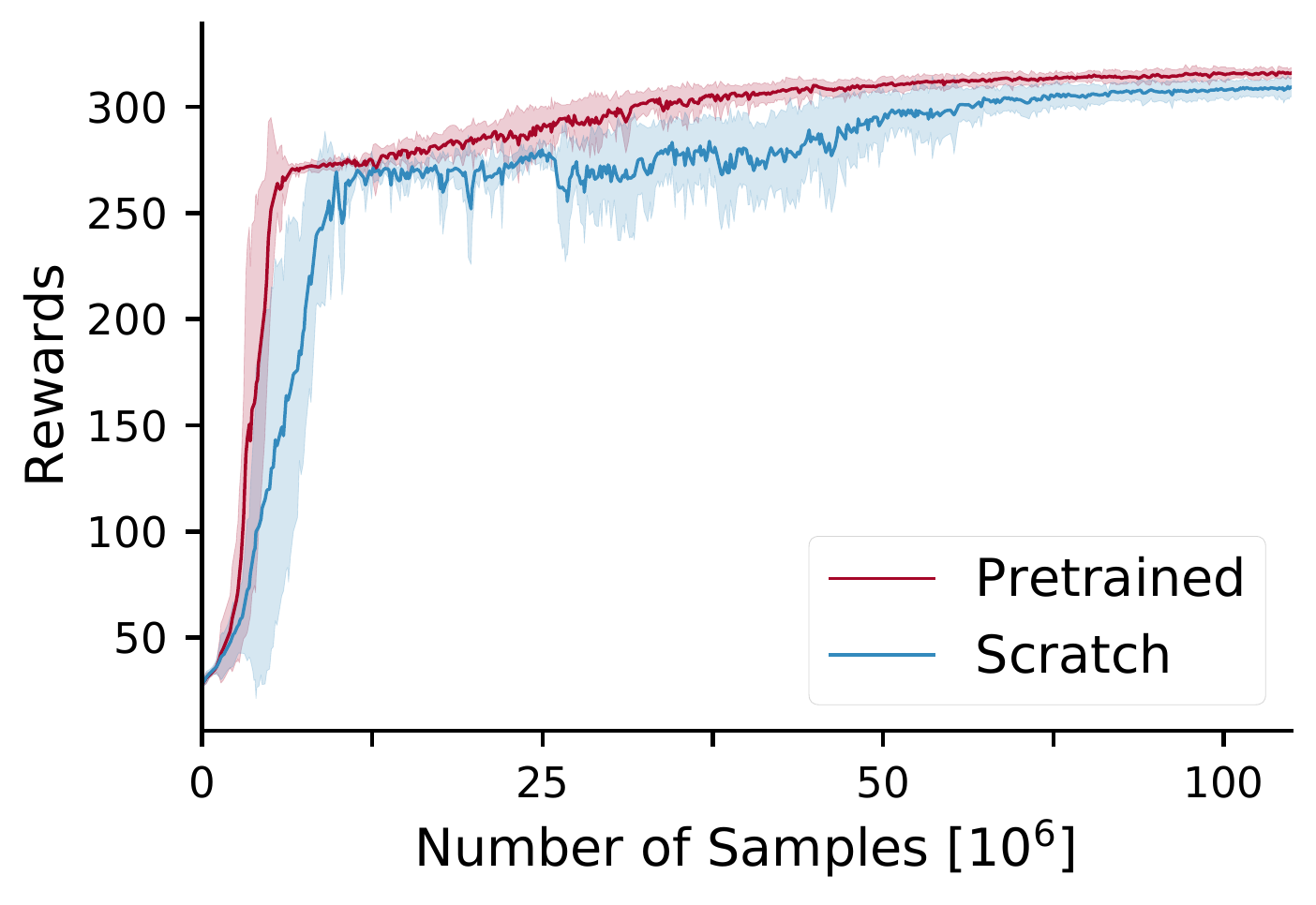}
         \caption{Training curve.}
         \label{subfig:samp_compare}
    \end{subfigure}
    \begin{subfigure}{0.28\textwidth}
         \centering
         \resizebox{0.85\columnwidth}{!}{%
          \begin{tabular}[b]{l|l|l}
            \hline
            Pattern & \thead{Avg Error[m]\\with Pretrain} & \thead{Avg Error[m]\\wout Pretrain}\\ \hline
             A         & 0.11$\pm$0.07  & 0.21$\pm$0.04      \\ \hline
             B         & 0.15$\pm$0.06  & 0.26$\pm$0.08      \\ \hline
             C         & 0.16$\pm$0.11  & 0.31$\pm$0.18      \\ \hline
             D         & 0.19$\pm$0.11  & 0.29$\pm$0.18      \\ \hline
             E         & 0.20$\pm$0.08  & 0.31$\pm$0.11      \\ \hline
             F         & 0.11$\pm$0.05  & 0.26$\pm$0.06      \\ \hline
             G         & 0.11$\pm$0.07  & 0.28$\pm$0.06      \\ \hline
             H         & 0.12$\pm$0.06  & 0.26$\pm$0.06      \\ \hline
            \end{tabular}
        }
        \label{table:training_compare_lstm}
        \caption{Performance comparison of training method in the middle of training process.}
    \end{subfigure}
    \begin{subfigure}{0.36\textwidth}
         \centering
         \includegraphics[width=\textwidth]{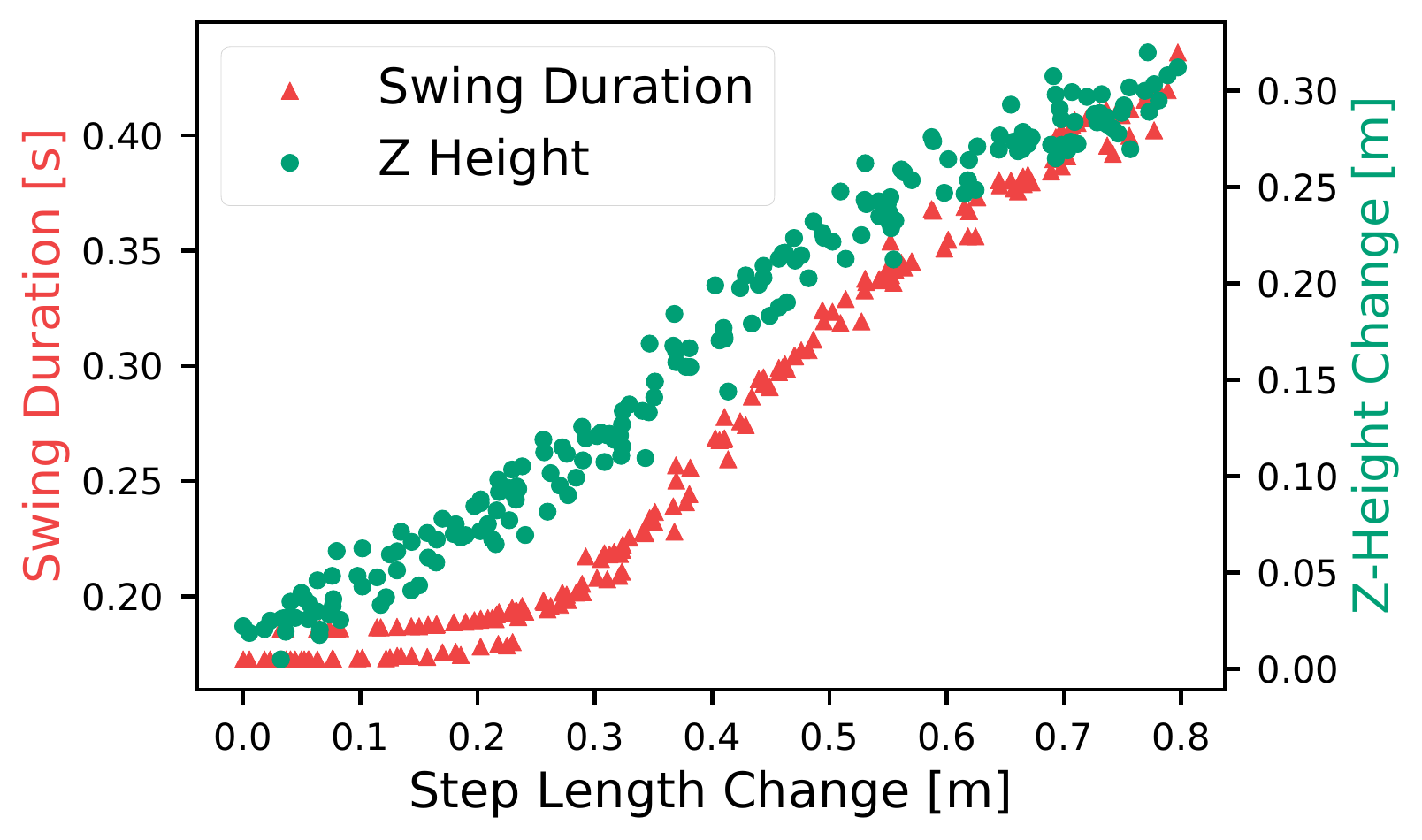}
         \caption{Adaptive swing duration.}
         \label{subfig:swing_time}
    \end{subfigure}
\caption{(a) The training curve shows the benefit of using a pretrained dynamics layer, which results in higher reward and faster convergence. (b) The performance of the policy is evaluated in during training ($\sim$50millions) shows that the pretrained method performs better than trained from scratch on the set of patterns. (c) The policy can only see the immediate next footstep command. As part of the emergent behaviors, the learned step frequency allows the robot to successfully achieve the next target step by taking a longer or shorter swing duration. The policy also learns to elevate the body height in order to enable longer steps.}
\label{fig:policy_result}
\vspace{-.3cm}
\end{figure*}

We compare the footstep performance of the bootstrapped policy versus a policy trained from scratch given the same amount of training experience. 
Figure \ref{subfig:samp_compare} shows that the policy using the pretrained dynamics module has faster convergence to a higher reward. 
As shown in Figure \ref{fig:policy_result}b, given only 50 million samples, the bootstrap method already shows twice the accuracy over the policy trained from scratch evaluated on all the patterns. Figure \ref{subfig:swing_time} shows the emergent behavior of the learned step frequency when the step length changes from shorter to longer.

\begin{table}[]
\vspace{4pt}
\centering
\resizebox{0.85\columnwidth}{!}{%
\begin{tabular}{l|l|l|l|l}
\hline
Pattern & \thead{Num\\Steps} & \thead{Avg\\Error[m]} & \thead{Max\\Error[m]}& \thead{Avg\\Speed[m/s]}\\ \hline
 A         & 3  & 0.06$\pm$0.02 & 0.10 & 0.97     \\ \hline
 B         & 4  & 0.09$\pm$0.03 & 0.18 & 1.23     \\ \hline
 C         & 4  & 0.10$\pm$0.02 & 0.18 & 1.42     \\ \hline
 D         & 4  & 0.08$\pm$0.05 & 0.22 & 1.36     \\ \hline
 E         & 4  & 0.11$\pm$0.04 & 0.22 & 1.39     \\ \hline
 F         & 6  & 0.08$\pm$0.01 & 0.12 & 1.26     \\ \hline
 G         & 7  & 0.08$\pm$0.02 & 0.13 & 1.37     \\ \hline
 H         & 8  & 0.07$\pm$0.01 & 0.10 & 1.31     \\ \hline
\end{tabular}}
\caption{This table shows the footstep accuracy results of the learned policy evaluated on each of the stepping stone pattern in simulation. 
Errors are measured between each footstep target to the center of the foot at the corresponding touchdown. We collect 100 trials in simulation for each pattern. The max error is the worst case among all trials for each pattern, which usually occurs at the last footstep for each pattern.
}
\label{table:sim_footstep}
\end{table}

\subsection{Simulation Performance on Stepping Stone Patterns}
Given the set of stepping stone patterns, we evaluate the learned policy in simulation. Table \ref{table:sim_footstep} shows the quantitative results from simulation. 
We note that the learned policy shows good performance on patterns A/F/G/H, as these patterns are more uniformly spread out. The performance over patterns B/C/D/E is degraded, since they have large changes in step length and direction between footsteps. This is more difficult for the policy to deal with, particularly since the policy only knows the immediate next footstep. 

\begin{figure}[h]
	\centering
    \begin{subfigure}{0.44\columnwidth}
         \centering
         \includegraphics[width=0.8\columnwidth]{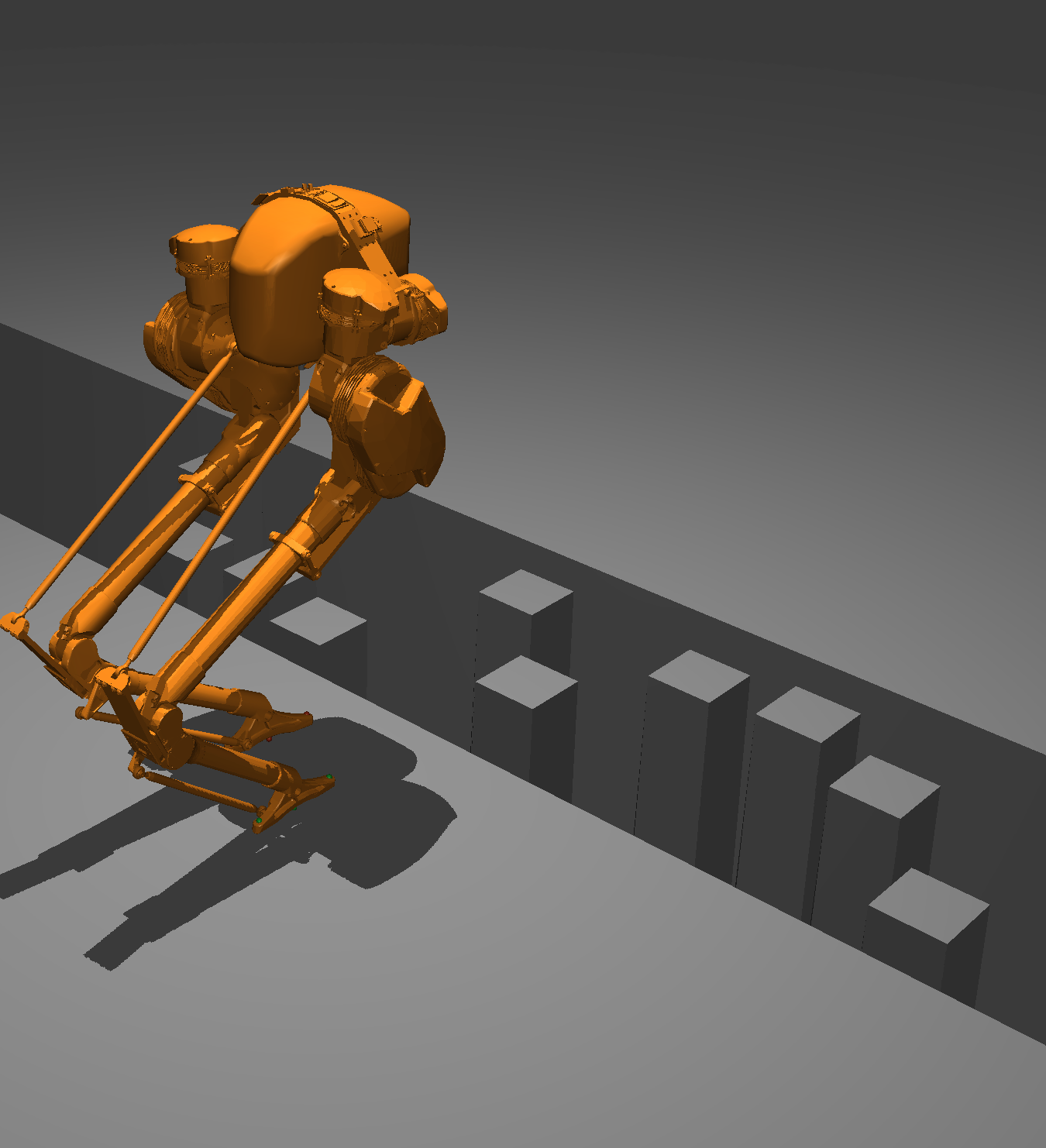}
         \label{subfig:sim_prediction}
    \end{subfigure}
    \begin{subfigure}[t]{0.54\columnwidth}
         \centering
         \resizebox{0.99\columnwidth}{!}{%
          \begin{tabular}{l|l}
            \hline
            \thead{Selection\\Method}     & \thead{Success\\Rate [\%]} \\ \hline
             Random              & 59.1     \\ \hline
             Random on TD side   & 69.0     \\ \hline
             Closest stone       & 50.2     \\ \hline
             Closest on TD side  & 76.8     \\ \hline
             Reachability model  & 82.3     \\ \hline
            \end{tabular}
        }
        \label{subtable:sim_prediction_test}
    \end{subfigure}
\caption{Simulation test of the utility of prediction model. \textbf{Left:} Stepping stone locations and initial robot position are randomized for each trial. Cassie is asked to go across the gap by stepping onto one of the stones. \textbf{Right:} ``Random" refers to random selection among the choices. ``Closest" and ``Closest on touchdown (TD) side" refer to selecting the closest (in euclidean sense) stone or only selecting on stones same as the touchdown side. The success rate is measured from 1000 independent simulation trials. Having a reachability model allows Cassie have the highest success rate of successfully crossing the gap.}
\label{fig:sim_prediction}
\end{figure}

\subsection{Simulation Evaluation of the Prediction Model}
\label{sec:pred_model_results}
\subsubsection{Model Accuracy}
As explained in section \ref{sec:pred_data_collect}, there are several choices for how we could collect data to train the footstep prediction model. 
We use a separate test dataset to evaluate which data collection method is better in terms of the corresponding trained model's accuracy. 
As shown  in  the  first  column  “Mean  error”  of  Table  II,  noisy procedural footsteps produces the best accuracy among the four methods.
It is interesting to note how bad dynamics randomization performs. This is because for similar touchdown states, we can have greatly different step error values due to different dynamics across experiments. This noise in the data may result in conflicting information about what the predicted step error should be, which reduces the accuracy of the model.

\subsubsection{1-Footstep Targeting with Prediction Model}
\label{sec:targeting1}

We also evaluate the prediction models with the learned footstep policy to see if they can be effectively used for single step reaction.
In simulation, the evaluation task asks the robot go across a 40cm wide gap with a single randomly placed stepping stone of size $10\times10\text{cm}^2$ in the gap. Note that Cassie's foot is 18cm long and 2cm wide. The stone can either be on the robot's left or right side based on which side is about to touchdown. During evaluation, Cassie walks towards the gap and as soon as the predicted error is less than a threshold ($<10$cm), the policy takes in the footstep commands towards the stepping stone and crosses the gap. A success means the robot is able to land on the stone safely and go across the gap without getting tripped or falling into the gap. Table \ref{table:model_prediction_test} lists the success rate for four types of trained models.
We find that the noisy procedural steps collection method results in the best overall model since it has nearly the best success rate for crossing the gap as well as the lowest prediction error. As such, we use this model for the remaining analysis. 

\begin{table}[]
\vspace{4pt}
\centering
\resizebox{0.9\columnwidth}{!}{%
\begin{tabular}{l|l|l}
\hline
Data collection method  & \thead{Avg\\error [cm]} & \thead{Success\\rate [\%]}\\ \hline
 Vanilla / Base            & 6.6  & 83.6     \\ \hline
 Dynamics randomization    & 9.1  & 73.7     \\ \hline
 Noisy Procedural steps    & 6.1  & 83.1     \\ \hline
 Noisy touchdown states    & 6.4  & 81.9     \\ \hline
 
\end{tabular}}
\caption{Each model is trained with the same hyper-parameters values for the same number of iterations. 
The ``Mean error" column shows the prediction accuracy on the evaluation data for reachability models trained with each training data collection method. The ``Success rate" column shows the 1-footstep targeting success rate over 1000 independent trials for each data collection method.
}
\label{table:model_prediction_test}
\end{table}

\subsubsection{Prediction Model over Multiple Choices}
We increase the gap size to 60cm and randomly place 10 stepping stones instead of just a single stone (Figure \ref{fig:sim_prediction}). The robot is initialized with different step commands and has to go across the gap by choosing one stone among the 10 possible choices that is predicted to have the smallest error. 
It should be noted that since the locations of the stones are random, there is no guarantee the majority or any of them will be good choices with prediction error less than 0.1m. 
Figure \ref{fig:sim_prediction} reports the success rate of each method over 1000 trials using our learned reachability model as well as four other heuristics (see caption) for selecting a stone. The reachability model achieved the highest success rate and significantly outperformed the closest-stone and random-stone heuristics.

\section{Hardware Results}
In this section, we consider the sim-to-real transfer capabilities using the Cassie robot with the overhead camera for perception. 



\subsection{Hardware Performance on Stepping Stone Patterns}
We test the same stepping stone patterns on hardware as we used in simulation.
During hardware execution, we manually switch the mode of operation between \emph{user-control} and \emph{footstep-reaction}. During \emph{user-control} mode, an operator is manually commanding footstep length and direction to bring the robot close to and facing the marker footstep pattern. We then enable the \emph{footstep-reaction} mode at an arbitrary time when the user decides the robot is in a ``good" initial state. Then, the policy will react to a sequence of footstep markers. Once the robot completes the number of footsteps in each pattern, the controller is switched back to \emph{user-control} and steps in place. 

\begin{figure*}[t!]
\vspace{4pt}
\centering
  \includegraphics[width=0.99\textwidth,angle=0]{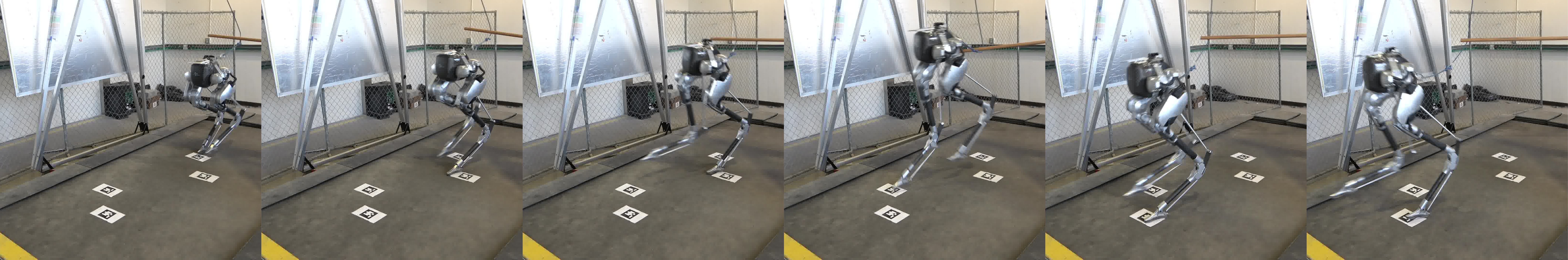}
  \caption{Cassie performs a sequence of footsteps (pattern D), which are shown as the markers (10x10cm) on treadmill.}
  \label{fig:motion_frames}
  \vspace{-.5cm}
\end{figure*}

We show that Cassie is able to exhibit the expected behaviors for all the patterns as shown in Figure \ref{fig:motion_frames}. Please refer to the accompanied video for hardware demonstrations. During hardware experiments, we noticed that it is difficult to get the policy to consistently hit all the markers every trial. We found that the robot initial state before enabling \emph{footstep-reaction} mode significantly affects the performance on hardware. For example, the initial speed and orientation of the robot affects the footstep accuracy, because the momentum carried before switching to \emph{footstep-reaction} mode influences the robot dynamics. This can be addressed with better orientation control of the policy and the use of look-ahead from a prediction model in future work.  

\subsection{Evaluation of the Prediction Model on Hardware}
\label{sec:targeting2}

We conducted two experiments on hardware similar to those in simulation. 
First, for 1-footstep targeting, we use the learned footstep prediction model to automatically choose when to switch to ``footstep-reaction" mode. Given the next footstep marker, the prediction model will tell the policy when the robot is in a state that will produce low footstep error, and then switch to ``footstep-reaction" mode. Please refer to the video submission for better illustration of the experiments.

To do this, the prediction model is called on every touchdown (based on the clock value) to check if the target marker has step error less than 0.1m. Once the condition is satisfied, the policy will send the target marker's relative location as the command and perform the subsequent footstep targeting. We found that 1-footstep targeting using the prediction model tends to be more consistent than user manually control the switch of the mode. 

Our second evaluation tests the prediction model on multiple footstep choices, similar to the tests performed in simulation. Instead of having a single marker for each footstep decision, we test having a set of possible targets formed by a stripe line. For each stripe, the prediction model chooses a reachable footstep from the given choices (20 choices are present along the line), indicated by the lowest prediction error. Then the policy controls the robot to land around the stripe line. We gradually increase the difficulty by stacking more stripe lines. As can be seen in the hardware video, Cassie is able to go across test cases with 1 or 2 stripe lines, but struggles when more lines are added. We hypothesize this challenge can be resolved by introducing look-ahead from the prediction model, so the initial state when approaching the stripe lines is better suited for dynamics evolving over such terrain. 

\section{Conclusion and Future Work} 
In this work, we demonstrate real-world 3D dynamic walking over stepping stones on a bipedal robot by introducing a learned 1-footstep control policy and a reachability prediction model. We examined the performance of the policy and the utilities of the prediction model both in simulation and on hardware. These two major pieces could form the foundations of future work, enabling dynamic locomotion with a multi-footstep look-ahead. 

Future works could focus on several directions. 
First, a reduced-order model planner that outputs footstep and body orientation commands could be incorporated into the learning process. This planner could be used during sampling so we could avoid randomly sampling commands which might be infeasible or out of scope for the desired task.
This simple model could also be used online with the policy to form a control hierarchy which keeps the planning process simple, but also has the robustness of a learned solution. 
Second, we seek to extend the learned prediction model with a look-ahead feature, so the model can better utilize the predicted next latent state recursively. Third, we hope to attach the camera to the robot for a true ego-centric view, allowing for maneuvers with self-contained on-board system. This would likely require a separate learning process to transform the camera information into the footstep commands. 

\addtolength{\textheight}{-8cm}   



\def\bibfont{\footnotesize}
\bibliographystyle{IEEEtranN}
\bibliography{main}

\end{document}